\def\year{2019}\relax

\documentclass[letterpaper]{article} %DO NOT CHANGE THIS
\usepackage{aaai19}  %Required
\usepackage{times}  %Required
\usepackage{helvet}  %Required
\usepackage{courier}  %Required
\usepackage[hyphens]{url}  %Required
\usepackage{graphicx}  %Required
\usepackage{algorithm}
\usepackage{algorithmic}
\usepackage{amsthm}
\usepackage{amssymb}
\usepackage{multicol}
\usepackage{amsmath}

\usepackage[font=small]{caption}

\usepackage{color}
\usepackage[textsize=scriptsize]{todonotes}
\definecolor{blue}{RGB}{0, 93, 170}			%Go Big Blue!
\definecolor{darkgreen}{RGB}{0, 102, 0}
\definecolor{orange}{RGB}{255, 160, 0} %'Cuse is in the house

\newcommand{\ignore}[1]{}
\usepackage[font=small]{subcaption}
\usepackage[font=small]{caption}

\newcommand{\citet}[1]{\citeauthor{#1}~\shortcite{#1}}
\newcommand{\citep}{\cite}
\usepackage{xspace}

\nocopyright
\title{Building Ethically Bounded AI}
\author{Francesca Rossi \\IBM Research and University of Padova \\ Yorktown Heights, NY, USA \\ francesca.rossi2@ibm.com
\And Nicholas Mattei \\ IBM Research \\ Yorktown Heights, NY, USA \\ n.mattei@ibm.com}

\begin{document}

\maketitle

\begin{abstract}
The more AI agents are deployed in scenarios with possibly unexpected situations, the more they need to be flexible, adaptive, and creative in achieving the goal we have given them. Thus, a certain level of freedom to choose the best path to the goal is inherent in making AI robust and flexible enough. 
At the same time, however, the pervasive deployment of AI in our life, whether AI is autonomous or collaborating with humans, raises several ethical challenges. AI agents should be aware and follow appropriate ethical principles and should thus exhibit properties such as fairness or other virtues.
%\nick{same as above, missing the end of a thought here}
These ethical principles should define the boundaries of AI's freedom and creativity. 
However, it is still a challenge to understand how to specify and reason with ethical boundaries in AI agents and how to combine them appropriately with subjective preferences and goal specifications. Some initial attempts employ either a data-driven example-based approach for both, or a symbolic rule-based approach for both. We envision a modular approach where any AI technique can be used for any of these essential ingredients in decision making or decision support systems, paired with a contextual approach to define their combination and relative weight. 
In a world where neither humans nor AI systems work in isolation, but are tightly interconnected, e.g., the Internet of Things, we also envision a compositional approach to building ethically bounded AI, where the ethical properties of each component can be fruitfully exploited to derive those of the overall system.
In this paper we define and motivate the notion of ethically-bounded AI, we describe two concrete examples, and we outline some outstanding challenges.
\end{abstract}

\section{Motivation and Overall Vision}

Whatever we do in our everyday life, be it at work or in our personal activities, we need to make decisions: what to eat, where to go on vacation, what car to buy, which route to take to go to work, what job to choose, and many more.
%Some of these decision are made in isolation, without any consultation with other individuals. But most of them are collective decisions, that we make together with others.
To make these decisions, we usually rely on our subjective preferences over the possible options. If we need to buy a car, we may have preferences over its color, its maker, its engine, and many other features. If we need to decide which restaurant to go for dinner, we may have preferences over location, facilities, food, drinks, and many other features.
However, subjective preferences are not the only source of guidance when making our decisions.
In many domains preferences are combined with moral values, ethical principles, or behavioral constraints that are applicable to the decision scenario and are \emph{prioritized} over the preferences \cite{Ro16,greene2014}. We have have our own preferences over food, but maybe the doctor recommended that we follow a diet to avoid some health issues, so we need to combine the doctor's guidelines with our taste preferences \cite{BaBoMaRo18,balakrishnan2018incorporating}. 
This is especially true in decision that may have an impact on others. In this context, social norms, regulations and laws could provide guidelines to follow when making a decision \cite{Sen,Thomson}. While driving our car, we may want to drive as fast as possible to get home sooner, but social norms and laws provide limits to speed and dangerous deriving behavior. 

AI systems are increasingly supporting human decision making, or they make decisions autonomously. So it is natural to ask ourselves how to code both subjective preferences and ethical principles in these systems.
This is especially necessary when AI systems tackle ill-defined problems whose solution procedure cannot be accurately defined by a rule-based approach but require data-driven and/or learning approaches, which are increasingly used in AI.
%
%When trying to inject subjective preferences and ethical priorities in a machine, it is important to be able to model these concepts, reason with them, and combine them, while at the same time keeping them separate to possibly give them different weights \cite{embed}.
%
Data-driven AI systems are indeed very successful in terms of accuracy and flexibility, and they can be very ``creative'' in achieving a goal, finding paths to the goal that could positively surprise humans and teach them innovative ways to solve a problem, such as the move that the AlphaGo system used against Lee Sedol in the 2017 match \cite{silver2017mastering} and a similar system that used uncommon methods to set records in Atari games \cite{Mnih2013}.
However, creativity and freedom without boundaries can sometimes lead to undesired actions: 
%but Concerns about the ways in which data-driven decision making systems behave when deployed in the real world are growing: what various stakeholder are worried about is that 
the system could achieve its goal in ways that are not considered acceptable according to values and norms of the impacted community. 

Recently researchers at DeepMind  collected a list of examples of ``specification gaming'' behaviors\footnote{More examples are available at: \url{https://vkrakovna.wordpress.com/2018/04/02/specification-gaming-examples-in-ai/}} and released AI Safety Block Worlds to examine these behaviors \cite{leike2017ai}. 
Examples of specification gaming includes:
\begin{itemize}
\setlength\itemsep{0em}
    \item 
a reinforcement learning agent in a boat racing game going in circles and repeatedly hitting the same reward targets in order to increase the score, instead of actually playing the game;
\item
a Eurisko game-playing agent that got more points by falsely inserting its name as the creator of high-value items;
\item
a Lego stacking system that flips the block instead of lifting, since lifting encouragement is implemented by  rewarding the z-coordinate of the bottom face of the block;
\item
a sorting program that always outputs an empty list, since it is considered a sorted list by the evaluation metric;
\item
a game-playing agent that kills itself at the end of level 1 to avoid losing in level 2;
\item 
a robot hand that pretends to grasp an object by moving between the camera and the object;
\item
a game-playing agent that pauses the game indefinitely to avoid losing.
\end{itemize}

The overriding concern is that the autonomous agents we construct may not obey some underspecified yet expected values on their way to maximizing some objective function \cite{simonite2018}. Thus, there is a growing need to understand how to constrain the actions of an AI system by providing 
boundaries within which the system must operate.

% As mentioned above, humans define their own boundaries via ethical principles, social norms, and laws. 
In bounding the behavior of AI systems, we may take inspiration from humans, who often constrain their decisions and actions according to a number of exogenous priorities, be they moral, ethical, religious, or business values \cite{Sen}, and we may want the systems we build to be restricted in their actions by similar principles \cite{conf/aaai/ArnoldKS17}. 
But how do we specify both subjective preferences and ethical boundaries in a machine? And how do we decide the relative weight for each of these two driving guidelines in making decisions? 

As for the ethical guidelines, the idea of teaching machines right from wrong has become an important research topic in both AI \cite{yu2018building} and in other disciplines \cite{wallach2008moral}. Much of the research at the intersection of AI and ethics falls under the heading of \emph{machine ethics}, i.e., adding ethics and/or constraints to a particular system's decision making process \cite{anderson2011machine}. One popular principle to handle these issues is called \emph{value alignment}, i.e., the idea that an agent can only pursue goals that follow values that are aligned to the human values and thus beneficial to humans \cite{russell2015research}. 
More generally, in the machine ethics field, the literature mentions both a so-called bottom-up approach, 
%A popular technique is called the \emph{bottom-up approach}, 
i.e., teaching a machine what is right and wrong by example \cite{allen2005artificial}, and a top-down approach, where explicit behavioral rules are specified,
as well as a combination of the two approaches.

For the subjective preferences,
since decision making is such a central task in AI systems, the study of how to represent \cite{RVW11a}, learn \cite{FuHu10a}, and reason \cite{DHKP11a,PTV15a} with preferences has been extremely active both within and beyond the field of AI \cite{goldsmith2009preference}, with significant theoretical and practical results \cite{DHKP11a,PTV15a} as well as libraries and datasets \cite{MaWa13a,MaWa17}. In many scenarios including multi-agent systems \cite{ShLe08a} and recommender systems \cite{RRSK11a}, user preference play a key role in driving the decisions the system makes. Thus AI researchers have defined many
preference modeling frameworks that allow for expressive and compact representations, effective elicitation techniques, and efficient reasoning and aggregation algorithms. 

Existing approaches to build ethical AI systems employ both data-driven or rule-based approaches. In the following section we will briefly describe two of them, to make the discussion more concrete. But many outstanding questions remain that we must address as a field. 

First, most approaches (like the two we will describe) use the same formalism for both the preferences and the ethical boundaries. This makes things easier, since priorities of the two kinds can be better compared and combined. 
However, it is important to allow for the possibility of a mixed approach. We may have rules describing the ethical boundaries but the agent's goal may need a data-driven approach, or vice-verse. In this generalized setting, it is not yet clear how to combine preferences and ethical boundaries, how to compare them, and how to combine them.

Second, most approaches try to design a single autonomous AI agent working in isolation. However, AI agents will increasingly work together with humans. It is not yet clear how to fruitfully split the task of achieving a goal while following ethical priorities in a team, rather than a single person or AI agent?  
Also, how can we link the ethical behavior of an AI system when it is composed of many sub-components, even if we can assure that each sub-component behaves within its ethical boundaries? This is increasingly relevant in IoT environments, where some certainty on the ethical properties of the overall system is necessary to trust, and thus adopt, the overall IoT system. 

Third, what ethical principles should be injected into AI systems? The same that humans use, or others?
How do we address the various cultural and temporal dynamics of the broad spectrum of human values and ethics?

The final point we would like to make is the role of the scientific associations, such as AAAI, to help resolve some of these questions, by adopting a multi-disciplinary and multi-stakeholder approach within their research community. 

\section{Two Examples of Existing Approaches}

Some initial attempts to build AI systems that obey both preferences (or some other optimization objective) and ethical guidelines
employ either a data-driven example-based approach for both, or a symbolic rule-based approach for both.

\subsection{A Symbolic and Logic-based Approach: Using CP-nets to Model Both Preferences and Ethical Priorities}

Preferences have been studied for many years within AI, and several formalism have been developed to model and reason with subjective preferences.
Each formalism has different different properties, related to compactness, expressive power, elicitation and learning, and reasoning efficiency.
Since ethical principles define the same kind of structures as preferences, that is, priority orderings over the possible decisions \cite{AllenVZ00,moral}, it seems reasonable to conjecture that ethical boundaries and priorities could be modeled using a (possibly adapted) existing preference frameworks.

This is the approach taken by \citet{LoMaRoVe18a},
%\nick{This is the citation to the AIES paper -- if you want it to be the bookchapter use LoMaRoVe18a} 
where the preference framework using CP-nets is used to also model and reason with ethical principles.
Among several existing preference representation languages described in the literature \cite{ABGP16a}, 
CP-nets \cite{cpnets} provide a qualitative way to compactly model preferences. They allow to express preferences over complex decisions made of several features, by stating contextual preferences over the values of each feature. For example, 
if we are choosing a car, we may prefer certain colors over others, and we may prefer certain makes over others. We may also have conditional preferences, such as in preferring red cars if the car is a convertible.
%in we want to express preferences over cars, we may decide a car's relevant features are
%its color, its engine, its model, its maker, and its price, and 
%with CP-nets we can for example say that, if a car is a Ferrari, we prefer it to be red. So we are stating a preference over the color of the car in dependence of its model.

CP-nets are a sequence of conditional preference statements like this one, and have been used widely in the preference reasoning community \cite{RVW11a,CGGM+15a,CELM08a}.
Each (acyclic) CP-net induces a partial order over the possible actions/outcomes; in the car example above, an outcome would be a complete specification of a car. CP-nets provide a compact way to model preferences: if the context in the cp-statements does not involve too many features, the induced order is exponentially larger than the CP-net.

In \citet{LoMaRoVe18} the authors show how to use CP-nets modelling both subjective preferences and ethical principles, and also how to measure the deviation between these two guidelines. If a person's preferences suggest actions that are too unethical, the ethical boundary should kick in and suggest (or enforce) alternative actions that are ethical 
within a threshold. 
This is done by defining a notion of distance between two CP-nets that is computed efficiency by adopting an approximation of the ``ideal'' distance between the induced orders \citet{loreggia2018}.

More precisely, two CP-nets are used: one models the preferences, and the other models the ethical priorities. An agent can make decisions using its subjective preferences only if these preferences are \emph{close enough} to the ethical principles, where being \emph{close enough} depends on a threshold over the CP-net distance. If instead the preferences diverge too much from the ethical principles, we analyze the agent's preference ordering until we find a decision that is a \emph{satisfactory compromise} between the ethical principles and the user preferences. The compromise is defined by setting a second threshold over distances between decisions of the two CP-nets. The ability to precisely quantify the distance between subjective preferences and external priorities, provides a way to both recognize deviations from feasibility or ethical constraints, and to suggest more compliant decisions \cite{LoMaRoVe18,LoMaRoVe18a}.

This approach thus allows to model preferences and ethical priorities in the same framework while being able to distinguish between them, and this provides the ideal environment to compare them, measure deviations between them, and define appropriate ways to combine them. CP-nets are just a set of logical preference rules. However, they have restrictions on their expressive power. Can we generalize this approach to allow also for the use of more expressive logics to define either the preferences and/or the ethical principles?

\subsection{A Data-Driven Approach: Reinforcement Learning and Ethical Examples}

In the standard model of online decision settings, an agent works by selecting one out of several possible actions at each time-step, such as recommending a movie to a user, or proposing a treatment to a patient in a clinical trial. 
Usually each of these actions is associated with a context, e.g., a user profile, and a feedback signal, e.g., the reward or rating.
%, is only observed for the chosen option. 
%In these online decision settings, the agent must learn the inherent trade-off between exploration, which involves identifying and understanding the reward from an action, and exploitation, which means gathering as much reward as possible from an action.  

In \citet{balakrishnan2018incorporating} 
%\nick{I put the other paper on arxiv, this will work for now} 
the authors consider cases where the behavior of the online agent may need to be restricted, by laws, values, preferences, or ethical principles. %\cite{russell2015research}.  
Therefore they apply a set of \emph{behavioral constraints} to the agent that are independent of the reward function. For instance, a parent or guardian group may want a movie recommender system (the agent) to not recommend certain types of movies to children, even if the recommendation of such movies could lead to a high reward \cite{balakrishnan2018incorporating}. In clinical settings, a doctor may want its diagnosis support system to not recommend a drug that typically works because of patient quality of life considerations. 

To model this scenarios, the authors adopt the {\em contextual multi-armed bandit} problem setting,
where the agent observes a {\em feature vector}, or {\em context}, to use along with the rewards of the arms played in the past in order to choose an arm to play. Over time, the agent learns the relationship between contexts and rewards and selects the best arm \cite{MaryGP15,AgrawalG13}. %\cite{MaryGP15,villar2015multi}.  
%
%In the MAB setting there are $K$ \emph{arms}, each associated with a fixed but unknown reward probability distribution \cite{LR85,UCB}. At each time step, an agent plays an arm, i.e., recommends an item to a user, and receives a reward that follows the selected arm's probability distribution, independent of the previous actions. A popular generalization of MAB is the contextual multi-armed bandit (CMAB) problem where the agent observes a $d$-dimensional {\em feature vector}, or {\em context}, to use along with the rewards of the arms played in the past in order to choose an arm to play. Over time, the agent learns the relationship between contexts and rewards and select the best arm \cite{AgrawalG13}. 
%
To model the ethical boundaries, they assume the agent is given both positive and negative examples of the correct behaviors, provided by a teacher agent, and the online agent must learn and respect these boundaries in the later phases of decision making. As an example, a parent may give examples of movies that their children can watch (or that they cannot watch) when setting up a new movie account for them. In \citet{BaBoMaRo18} a graphical interface for this system is demonstrated as well as the effect on overall reward by imposing exogenous constraints.

Hence, the overall system learns two policies: a reward-based one and an ethical one. This approach allows for some flexibility in how much the ethical boundaries override the reward signal, i.e., the preferences of the user. This is done by exposing a parameter of the algorithm that allows the system designer to smoothly transition between the two policy extremes: the one where the agent is only following the learned constraints and is insensitive to the online reward, and the other extreme where the agent is only following the online rewards and not giving any weight to the learned ethical principles.  This work has been recently extended to a multi-step setting with reinforcement learning where multiple policies are blended together by a bandit-based orchestrator \cite{noothigattu2018interpretable}.

\section{Outstanding Challenges}

We have seen just two examples of how the current literature concretely addresses the problem of embedding ethics into AI systems; see the survey by \citet{yu2018building} for more. We chose these two examples as we have been directly involved in these efforts, but also because we see them as prototypical of two complementary approaches: the \emph{top-down} approach following symbolic and logic-based formalisms and the \emph{bottom-up} approach focused on data-driven machine learning techniques.

\smallskip
\noindent
\textbf{Combining Rule-Based and Data-Driven Approaches.}
Using the same approach for both the goal and preference specification  of the agent and the ethical boundaries makes things easier for those who design and implement these systems. Priorities expressed by both the preferences and ethics can be easily compared and combined if they are modelled with the same formalism. 
However, it is important to allow for the possibility of a mixed approach. We may have rules describing the ethical boundaries but the agent's goal may need a data-driven approach, or vice-verse. So it is important to understand how to combine and compare rule-based and logic-based approaches on one side, and data-driven machine learning approaches on the other. In this generalized setting, how do we measure deviation between objects of these two kinds? How do we decide what action should be taken when we realize the preferences to achieve the agent's goal are too far from the ethical guidelines?

\smallskip
\noindent
\textbf{AI/Humans Teams and IoT.} 
Most existing approaches aim to build autonomous AI agents, but in real life agents will increasingly work together with humans. Preferences and ethical principles apply to teams of agents and humans, but they are not necessarily the same for these two kinds of members in the team. For example, can AI play the role of advising and guiding humans to better follow ethical guidelines? How can we split the task of achieving a goal while following ethical priorities in a team, rather than a single person or AI agent?  In \citet{embed} an initial overall approach to embed ethical principles in collective decision making has been proposed. How do we go from that approach to concrete processes and mechanisms to build ethically bounded AI/humans teams?

When moving from single agents to teams of agents, it is also important to employ a compositional approach to proving the ethical properties of an AI system. The ideal situation is one where the composition of ethically bounded AI systems is also ethically bounded. The next best situation, probably much more realistic, is one where the ethical behavior of the components allow us to derive some information on the ethical behavior of the whole system (such as in \cite{SrRo18a}). Without some form of compositionality, it will be risky to combine many AI systems, such as done when constructing IoT systems, even if each one of the systems is ethically bounded, since we would not be able to trust the overall system in terms of its ethical properties.

%I will then propose a way to generalize and move forward from the current approaches, describing a vision of future ethically bounded AI that can also help humans identify and follow the appropriate ethical boundaries for them.
%\nick{them is ambigous?  do you mean the AI or the human here? the humans}

\smallskip
\noindent
\textbf{Who Decides the Ethical Boundary?}
Even assuming we understand how to build ethically bounded AI systems, who is going to decide the ethical principles to be injected into such systems? Are human values suitable also for machines, given that machines have extended capabilities compared to humans but also lack some very relevant human feelings, such as guilt or empathy, that heavily support human's ethical behavior?

What is ethical in one culture may not be considered ethical in another culture. How can we build AI systems that can be deployed globally and behave appropriately depending on where they will function? In addition, ethical principles changes over time. How can we build this evolving capability in ethically bounded AI system? Once deployed, how can an AI system itself, or a human using it, make sure that its ethical boundary evolves together with the surrounding human community?

%I will finish the talk by addressing meta-issues such as the identification of the suitable values to embed into ethical AI, the possible difference between human and AI ethics, the cultural differences around ethics, 

\smallskip
\noindent
\textbf{The Role of Scientific Associations.}
Scientific associations such as AAAI
can help societies and corporations to define and build ethically bounded AI. These associations represent research communities where the ideas first get discussed and reviewed by peers. However, these ideas, especially those that address societal issues such as the ethical boundary for AI systems, should also be discussed with experts of other disciplines, such as social scientists and economists. And such multi-disciplinary discussion should go in both directions: from AI to social sciences, to understand the impact of the proposed solutions to the society, and from social sciences to AI, to drive AI research to address the societal challenges we face through a pervasive use of AI.

A multi-disciplinary discussion is therefore necessary, but it is not sufficient. In addition, the impacted users and communities should have their voice heard. Consumer rights associations, civil society groups, comparative multi-cultural study groups, policy makers, should all be part of a wide educational and research effort that should aim to funnel technical solutions in the appropriate direction.

AAAI and other technical scientific associations should lead or at least be very active part of this multi-disciplinary and multi-stakeholder discussion,
hosting events and efforts within the research community
that can expose AI researchers to ideas and points of views from other disciplines and different stakeholders.  In addition, these organizations can create resources for both practitioners\footnote{\url{https://medium.com/design-ibm/everyday-ethics-for-artificial-intelligence-75e173a9d8e8}} and students \cite{GoKoKuMa17} to learn about AI ethics.

Existing efforts, such as the AIES conference and the 
AAAI 2019 track on AI for Society, as well as panels and invited talks on ethics for AI, e.g., Max Tegmark's IJCAI 2018 talk and Nick Bostrom's AAAI 2016 talk, are a good starting point, but they need to be followed by concrete initiatives to facilitate multi-disciplinary research and give value to studies on the impact of AI on society. All this can and should be done in concert with the many existing initiatives around beneficial AI, such as the Partnership on AI, the IEEE Ethics in Action initiative, the Future of Life Institute, the Center for the Future of Intelligence, and the many other academic labs and teams focusing on ethical and beneficial AI.

\end{document}